\documentclass[letterpaper]{article} % DO NOT CHANGE THIS
\usepackage{aaai25}  % DO NOT CHANGE THIS
\usepackage{times}  % DO NOT CHANGE THIS
\usepackage{helvet}  % DO NOT CHANGE THIS
\usepackage{courier}  % DO NOT CHANGE THIS
\usepackage[hyphens]{url}  % DO NOT CHANGE THIS
\usepackage{graphicx} % DO NOT CHANGE THIS
\usepackage{natbib}  % DO NOT CHANGE THIS AND DO NOT ADD ANY OPTIONS TO IT
\usepackage{caption} % DO NOT CHANGE THIS AND DO NOT ADD ANY OPTIONS TO IT
\usepackage{algorithm}
\usepackage{algorithmic}
\usepackage{subcaption}
\usepackage{amsmath}
\usepackage{amssymb}
\usepackage{dsfont}
\usepackage{mathrsfs}
\usepackage{newfloat}
\usepackage{listings}
\usepackage{times}
\usepackage{latexsym}
\usepackage[T1]{fontenc}
\usepackage[utf8]{inputenc}
\usepackage{microtype}
\usepackage{csquotes}
\usepackage{inconsolata}
\usepackage{graphicx}
\usepackage{xcolor}

\usepackage{framed}
\usepackage{multirow} % Include this package in your preamble
\usepackage{graphicx}
\usepackage{subcaption}  % For side-by-side figures with captions
\usepackage{comment}
\usepackage{xspace}
\usepackage{amsmath}
\newcommand{\update}[1]{#1}

\newcommand{\segsub}{\textsc{SegSub}\xspace}
\newcommand{\updatedlabel}{$l_{new}$}
\newcommand{\retlabel}{$l_{RET}$}
\newcommand{\ret}{$l_{RET}$}

\usepackage{ifthen}
\newboolean{arxiv}
\setboolean{arxiv}{false} % Change to false for AAAI submission

\ifthenelse{\boolean{arxiv}}{
  \usepackage[colorlinks=true, allcolors=blue]{hyperref}
}{
    \usepackage{xstring}
    
    \newcommand{\autoref}[1]{%
      \begingroup
      \def\reftext{}%
      \IfBeginWith{#1}{fig:}{\def\reftext{Figure~}}{}%
      \IfBeginWith{#1}{tab:}{\def\reftext{Table~}}{}%
      \IfBeginWith{#1}{sec:}{\def\reftext{Section~}}{}%
      \IfBeginWith{#1}{subsec:}{\def\reftext{Subsection~}}{}%
      \IfBeginWith{#1}{eq:}{\def\reftext{Equation~}}{}%
      \reftext\ref{#1}%
      \endgroup
    }
}

\usepackage{booktabs}
\usepackage{graphicx}
\usepackage{ifthen}
\usepackage{csquotes}

\title{\segsub: Evaluating Robustness to Knowledge Conflicts and Hallucinations in Vision-Language Models}

\author{Peter Carragher\thanks{\textbf{Correspondence:} petercarragher@cmu.edu}, Nikitha Rao, Abhinand Jha\thanks{Work done during affiliation with Carnegie Mellon University, now at Google.}, R Raghav, Kathleen M. Carley}
\affiliations{
  Carnegie Mellon University \\
  Pittsburgh, PA 15217
}

\begin{document}
\maketitle

\begin{abstract}
Vision language models (VLM) demonstrate sophisticated multimodal reasoning yet are prone to hallucination when confronted with knowledge conflicts, impeding their deployment in information-sensitive contexts. While existing research addresses robustness in unimodal models, the multimodal domain lacks systematic investigation of cross-modal knowledge conflicts. This research introduces \segsub, a framework for applying targeted image perturbations to investigate VLM resilience against knowledge conflicts. Our analysis reveals distinct vulnerability patterns: while VLMs are robust to parametric conflicts ($\sim20\%$ adherence rates), they exhibit significant weaknesses in identifying counterfactual conditions ($<30\%$ accuracy) and resolving source conflicts ($<1\%$ accuracy). Correlations between contextual richness and hallucination rate (r = -0.368, p = 0.003) reveal the kinds of images that are likely to cause hallucinations. Through targeted fine-tuning on our benchmark dataset, we demonstrate improvements in VLM knowledge conflict detection, establishing a foundation for developing hallucination-resilient multimodal systems in information-sensitive environments.

\end{abstract}

\begin{links}
    \link{Code}{https://github.com/CASOS-IDeaS-CMU/SebSub}
    \link{Datasets}{https://www.doi.org/10.1184/R1/28297076}
\end{links}

\section{Introduction} %1/2 - 1
The proliferation of VLMs has brought about significant advancements in multimodal reasoning capabilities, particularly within Visual Question Answering (VQA) systems that interpret image content and respond to text-based queries. However, this also raises concerns of AI-generated misinformation and VLM hallucinations. Underlying these concerns, the capacity of automated systems to reason over knowledge conflicts presents a challenge for information ecosystem integrity. 

The relationship between knowledge conflicts, hallucinations, and AI-generated misinformation has been well characterized by frameworks for information disorder \citep{wardle2017information} and LLM hallucination \citep{huang_survey_2025}. For example, empirical investigations have found that unimodal question answering systems are vulnerable to knowledge conflicts between encoded parametric knowledge and external contextual information sources \citep{neeman_disentqa_2022,longpre_entity-based_2022}. However, the multimodal research landscape \citep{liu2024visual} lacks methodologically rigorous investigation of cross-modal knowledge conflicts \citep{conflicts-main-survey}. 

Particularly in multimodal contexts where image manipulation is increasingly sophisticated \citep{tandoc_defining_2018}, the evaluation of model robustness under visual knowledge conflicts is well motivated. Accordingly, we propose a visual knowledge conflict taxonomy with three categories---source conflicts (between multiple contradictory image sources), parametric conflicts (between knowledge encoded in the model and image source), and counterfactual conflicts (between the query and image source, such that the query presupposes nonexistent visual elements). Grounded in existing LLM hallucination taxonomy \citep{huang_survey_2025}, parametric conflicts correspond to ``factual contradictions'' that can arise from over-fitting to training data, source conflicts correspond to ``logical inconsistencies''---hallucinations that represent failures in reasoning across multiple information sources---and counterfactual conflicts correspond to ``extrinsic hallucinations''--- the generated answer cannot be verified using the given image source as key visual elements have been removed.

\begin{figure}
    \centering
    \includegraphics[width=0.95\linewidth]{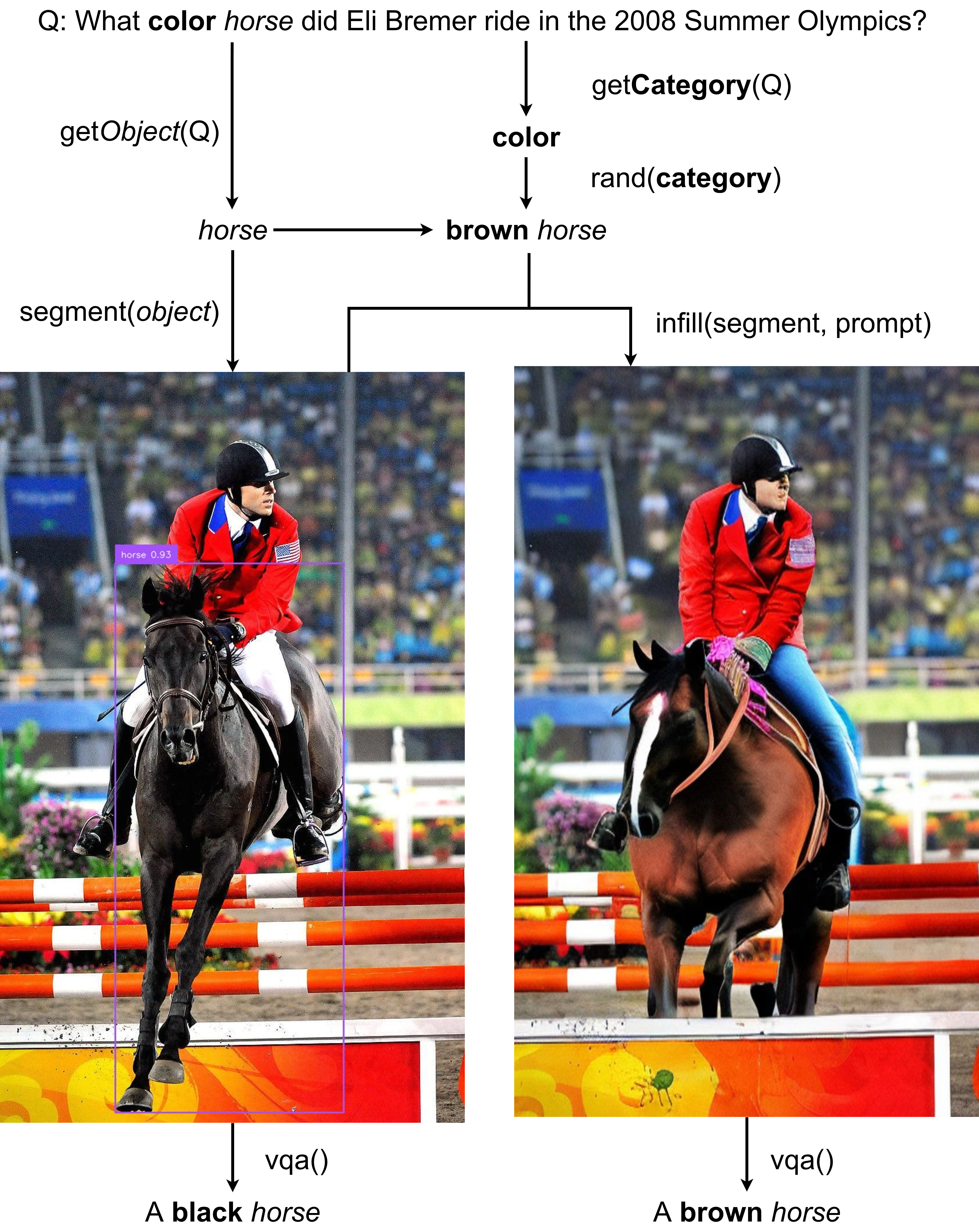}
    \vspace{-2mm}
    \caption{The \segsub framework: given a VQA task, we perturb existing (image, question, answer) triples with new images and answers to augment the dataset.}
     \vspace{-4mm}
    % d5c76d760dba11ecb1e81171463288e9
    \label{fig:seg_sub_pipeline}
\end{figure}

To operationalize this taxonomy, we introduce \segsub (\autoref{fig:seg_sub_pipeline}), a segmentation substitution framework for generating knowledge conflicts through image perturbations. 
% The framework implements three distinct augmentation strategies corresponding to each conflict classification. 
For source conflicts, we introduce contradictions between multiple information sources, creating scenarios that evaluate multi-source reasoning capabilities (\autoref{fig:perturbed_examples}). For parametric conflicts, we modify salient visual attributes of queried objects, creating measurable inconsistency between expected responses and visual evidence (as demonstrated in \autoref{fig:seg_sub_pipeline}). For counterfactual conflicts, we systematically remove queried objects while preserving contextual cues, creating ideal conditions for model hallucination (\autoref{fig:bat_boy}). Applying this framework to WebQA~\cite{chang_webqa_2021}, VQAv2~\cite{goyal2017making}, and OKVQA~\cite{marino_ok-vqa_2019}, we generate a benchmark dataset comprising over 35,000 systematically perturbed samples.

% The \segsub framework provides a systematic approach for evaluating the robustness of visual reasoning under various knowledge conflict conditions. In particular, 

The generated \segsub benchmark dataset can be used to measure susceptibility to hallucination. Our analysis yields several important findings regarding multimodal hallucination patterns in state-of-the-art systems. VLMs demonstrate relatively high robustness to parametric conflicts, with incorrect adherence to original labels occurring in only 20\% of perturbed samples (\autoref{fig:parametric_effect}). However, this raises concerns over the degree to which the reasoning of VLMs is vulnerable to image manipulation, and is in contrast to empirical research on unimodal systems, where LLMs rely on parametric knowledge when presented with conflicting textual sources \citep{hong_why_2024}. Accuracy rates are below 30\% for counterfactual conflicts and below 1\% for source conflicts. In the latter case, models entirely disregard contradictory evidence and attend to a single information source.

To overcome these reasoning failures, targeted fine-tuning on knowledge conflict data is applied, leading to improved performance on counterfactual conflicts across all datasets and models tested. Futhermore, we find a significant correlation between contextual richness and hallucination rate on counterfactual samples ($r = -0.368, p = 0.003$), motivating future research into using visual context cues as a predictor of hallucination risk in multimodal systems. Thus, the \segsub framework establishes practical detection and mitigation strategies for visual knowledge conflicts and hallucinations. Together, these contributions form a foundation for developing multimodal systems that are capable of maintaining factual consistency in complex information environments.

\section{Related Work} %2/3
% Our work stands at the intersection of several well-established pillars of research---knowledge conflicts in text sources \citep{conflicts-main-survey,conflicts-survey-1}, dataset augmentation for QA \citep{neeman_disentqa_2022,dataset-gen-attack,dataset-gen-contradoc,dataset-gen-contraqa}, VQA tasks \citep{marino_ok-vqa_2019,goyal2017making}, and diffusion models for text-to-image generation \citep{ravi_sam_2024,liu_grounding_2024,rombach_high-resolution_2022,yu_inpaint_2023,rombach_high-resolution_2022}.
% Our work stands at the intersection of several well-established pillars of research---knowledge conflicts in text sources, dataset augmentation for VQA tasks, and diffusion models for text-to-image generation.

Prior work on knowledge conflicts falls into two broad categories; the construction of evaluation datasets, and method-based contributions for model finetuning. Along these lines, \segsub uses conditional image generation and diffusion models to generate mulitmodal knowledge conflicts and finetune VLMs.

% \vspace{-2mm}
%\subsection{Augmenting Reasoning Tasks for LLM Evaluation \nr{rephrase text to avoid overflow}}
% \paragraph{Robustness of LLMs against reasoning tasks} 
% Furthermore, the reasoning required by VQA datasets is likely quite different than that required for arithmetic, logical reasoning, and algorithmic problems, and so the findings on these domains may not translate to the vision language setting. We aim to address this gap in the literature by applying image perturbations to augment VQA datasets for the purpose of VLM evaluation.
% \paragraph{Knowledge conflicts in QA tasks}
\paragraph{Knowledge Conflict Evaluation}
Recent work on evaluation has shown that LLMs are not robust to perturbations in text-based reasoning tasks \citep{zhang_darg_2024,mirzadeh_gsm-symbolic_2024,zhu2023dyval,wang2024benchmark} and that LLM performance degrades when conflicts exist in the source data for QA tasks \cite{conflicts-main-survey,conflicts-survey-1}. Longpre et al. \citep{longpre_entity-based_2022} introduced an entity-based knowledge conflict framework for evaluating how models handle conflicting information between learned parametric knowledge and contextual (non-parametric) data. Chen et al. \citep{chen_rich_2022} evaluate QA model on source conflicts. Hong et al. \citep{hong_why_2024} induce hallucinations in retrieval-augmented models by introducing counterfactual noise, which they define as conflicting but contextually relevant information. They also find that retrieval-augmented models ignore conflicting sources. 
% This bias can lead to reduced confidence and less accurate answers. Our work seeks to explore these issues in a multimodal context. 

\paragraph{Knowledge Conflict Fine-tuning}
Attempts to address this reasoning gap in LLMs include fine-tuning on both human annotated \cite{dataset-hum-contradict-wiki,dataset-hum-contradict-claim} and LLM generated \cite{dataset-gen-attack,dataset-gen-contradoc,dataset-gen-contraqa} datasets. Generative approaches involve extending a base dataset like SQuAD \cite{dataset-squad} to include sources with conflicting information \cite{controllable-memory}. DisentQA is formed by a combination of prompting and entity-substitution techniques \citep{neeman_disentqa_2022}. Recent work demonstrates that LLMs can be trained to retrieve more relevant context when the parametric information and provided sources are insufficient \citep{labruna2024retrieve,wang2024learningretrieveincontextexamples}. Our work addresses the gap between these foundations and multimodal QA systems \cite{conflicts-main-survey} by fine-tuning VLMs with knowledge conflicts to recognize when visual evidence is insufficient to complete the VQA task.

% Our work builds on this idea by applying similar principles to VQA datasets, allowing us to observe how Vision Language Models (VLMs) navigate visual knowledge conflicts.

% \paragraph{Datasets for evaluating knowledge conflicts} 

% , diminishes performance in scenarios requiring nuanced reasoning.  By fine-tuning models on datasets that include structured conflicts, our research aims to evaluate how effectively VLMs can disentangle and adapt to competing visual sources in VQA tasks.

% Existing work primarily focuses on only text-based knowledge conflicts whereas in this work, we propose a curated knowledge conflict dataset consisting of images. Following the approach in DyVal \cite{wang2024benchmark} we run a post-generation validation check for each our of generations to filter out noisy samples.

\paragraph{Conditional Image Generation}
Along with discriminative models that can segment images \citep{ravi_sam_2024,liu_grounding_2024}, advancements in Computer Vision have resulted in diffusion models that can generate images \citep{rombach_high-resolution_2022} based on textual prompts. Generative Adversarial Networks have proven successful in conditional generation \citep{lu_cigli_2021}, such as modifying the color of specific objects in an image \citep{khodadadeh2021automatic}. While naive approaches to counterfactual robustness include image masking \citep{chen2020counterfactual} and noising \citep{ishmam2024visual}, these recent advances enable a generative approach.

Counterfactual image generation has been used for several distinct tasks, from human AI teaching \citep{goyal_counterfactual_nodate} and object classification \citep{sauer_counterfactual_2021}, to model explainability \citep{vermeire_explainable_2022,chang_explaining_2019}. Overall, the focus is on image classifiers, how they are susceptible to noise, and how counterfactuals can help interpret the inner workings of these classifiers. As of yet, counterfactual image generation has not been used for inducing knowledge conflicts. In this work, we apply image segmentation \citep{yu_inpaint_2023,rombach_high-resolution_2022,suvorov_resolution-robust_2022} and conditional image generation to create counterfactual images by segmenting and then infilling or inpainting objects in an image. This method allows us to augment existing VQA datasets and finetune VLMs to enhance robustness against knowledge conflicts and counterfactual samples.

\paragraph{Detecting Manipulated Images}
Another long line of work focuses on detecting generated or perturbed images \citep{dong_think_2022,luo_lare2_2024,ojha_towards_2023}. The \segsub framework could be adopted for this task by making all generations positive samples. We discuss this as an avenue for future research.

% \paragraph{\todo{Counterfactual Image Generation}}

% \paragraph{Instruction-based Image Editing}

% \begin{enumerate}
%     \item \citep{goyal_counterfactual_nodate}: CF visual explanations. very different task from us
%     \item \citep{sauer_counterfactual_2021}: CF generative networks. "CFs improve out of distribution robustness, with marginal drop in original (image classification) task". We can use this to back up our findings, but our tasks/goal is different.
%     \item \citep{vermeire_explainable_2022}: Explaining image classifiers by CF generations: the infill generation is similar, can use it to back up our work
%     \item \citep{chang_explaining_2019}: similar to \citep{vermeire_explainable_2022}. 
% \end{enumerate}

% datasets and models for Instruction-based Image Editing, wherein models are evaluated on how faithfully they can follow text-based image editing prompts \citep{hui_hq-edit_2024,bodur_iedit_2023,zhang_magicbrush_2024}. The segmentation substitution method proposed in this work is parallel to these efforts and may be used to generate synthetic instruction-based image editing datasets similar to \cite{bodur_iedit_2023}. However, the focus of this paper will be on the application of these methods to the augmentation of visual reasoning tasks.
% % Lynnette's GAN \citep{lu_cigli_2021}

% \section{WebQA Task} %1/4
% \input{sections/03-task}

% \section{Model Description}
% \input{sections/04-model_description}

\section{Methodology} %1-3
\begin{table}[t]
    \caption{Distribution of the VQA datasets.}
    \vspace{-2mm}
    \label{tab:org_dataset}
    \centering
    \resizebox{\columnwidth}{!}{\begin{tabular}{lrr}
    \toprule
        \textbf{Dataset} & \textbf{\# Training samples} & \textbf{\# Validation samples} \\
        \midrule
        
        WebQA & 8634 & 1081 \\
        VQAv2 & 7765 & 1830 \\
        OK-VQA & 0 & 474 \\
        \midrule
        \segsub & 30155 & 5070 \\
        \midrule
        Total & 46554 & 8455\\
        \bottomrule
    \end{tabular}}
    \vspace{-4mm}
\end{table}

\segsub is a framework designed to enhance the robustness of VLMs by augmenting existing VQA datasets with the intention of introducing knowledge conflicts using perturbed images. Quality checks ensure that noisy perturbations are filtered out before we finetune models on the generations. Model performance is then evaluated on both the original and perturbed datasets. Finally, we analyze the effect of image-question contextualization on hallucination rate for counterfactual conflicts.

%Inspired by the Entity Replacement Framework \citep{longpre_entity-based_2022}, we term our framework for generating perturbed images Segmentation Substitution (SegSub). 

\subsection{The \segsub~Framework}
\label{sec:framework}
%Through targeted perturbations following the process outlined in \autoref{fig:seg_sub_pipeline}, the \segsub framework generates image samples on which existing VLMs struggle to reason. 
\autoref{fig:seg_sub_pipeline} gives an overview of the framework. First, given a QA pair with image sources $i_1, ..., i_n$, we prompt Gemini-1.5-flash to extract the noun that functions as the object of the question. We then prompt the Segment Anything Model v2 (SAMv2) \citep{ravi_sam_2024, liu_grounding_2024} to segment the object of the question in each image $i$. Finally, we apply a perturbation to the segmented regions by either removing the object from the image using Large Mask Inpainting (LaMa) \citep{suvorov_resolution-robust_2022} or changing the color or shape of the object using Stable Diffusion \citep{rombach_high-resolution_2022}. These perturbations generate three types of knowledge conflict.

\begin{figure}
    \centering
    %  [trim={left bottom right top},clip]
    \includegraphics[width=\linewidth,trim={0cm 4cm 10cm 0cm}]{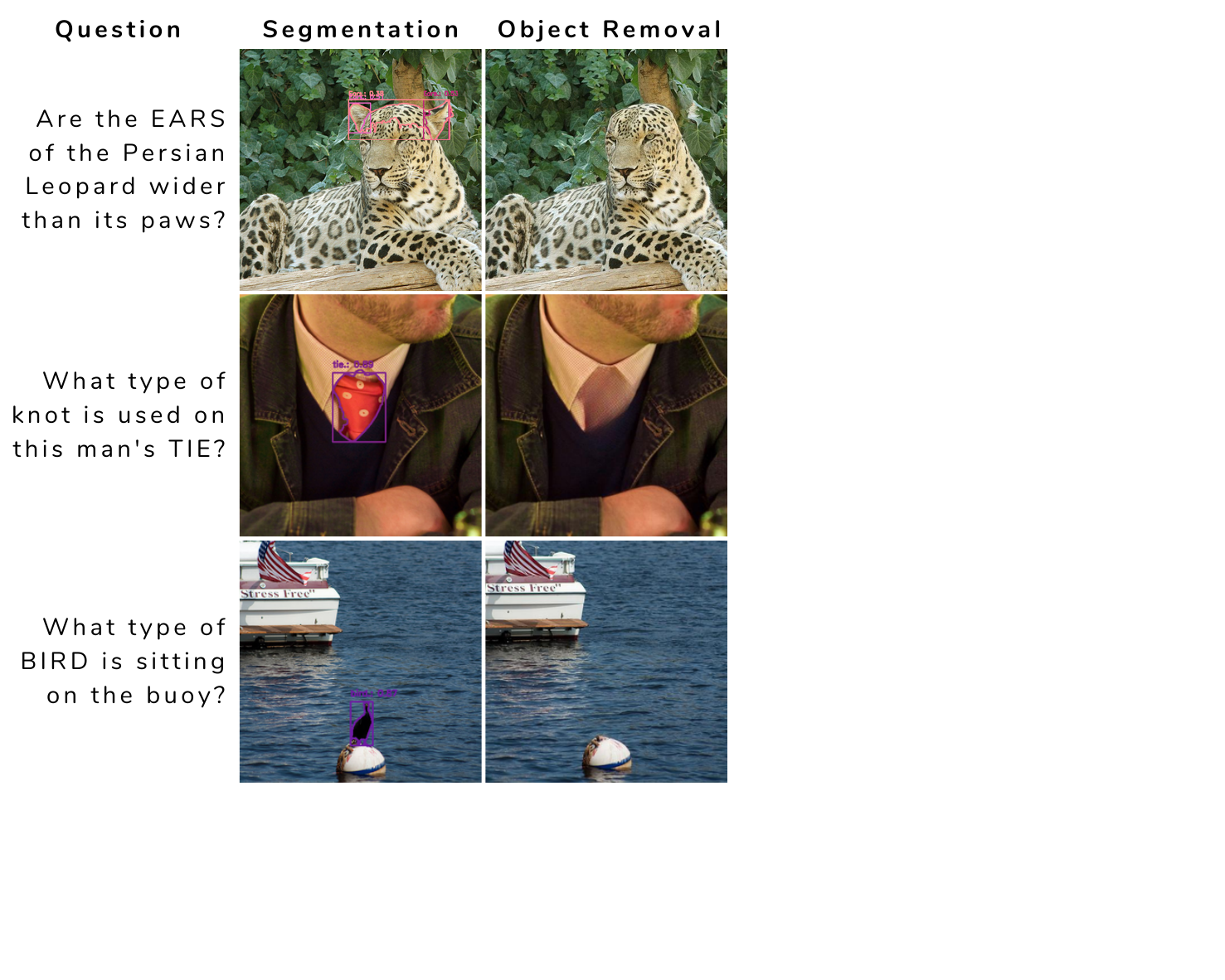}
    \caption{Examples of original images and counterfactual image generations. At the time of writing, ChatGPT hallucinates on these examples.}
    \label{fig:counterfactual_examples}
\end{figure}

\subsection{Knowledge Conflict Types}
\label{sec:tasks}

\begin{figure*}
    \centering
    \includegraphics[width=0.8\linewidth,trim={0cm 4cm 0cm 0cm}]{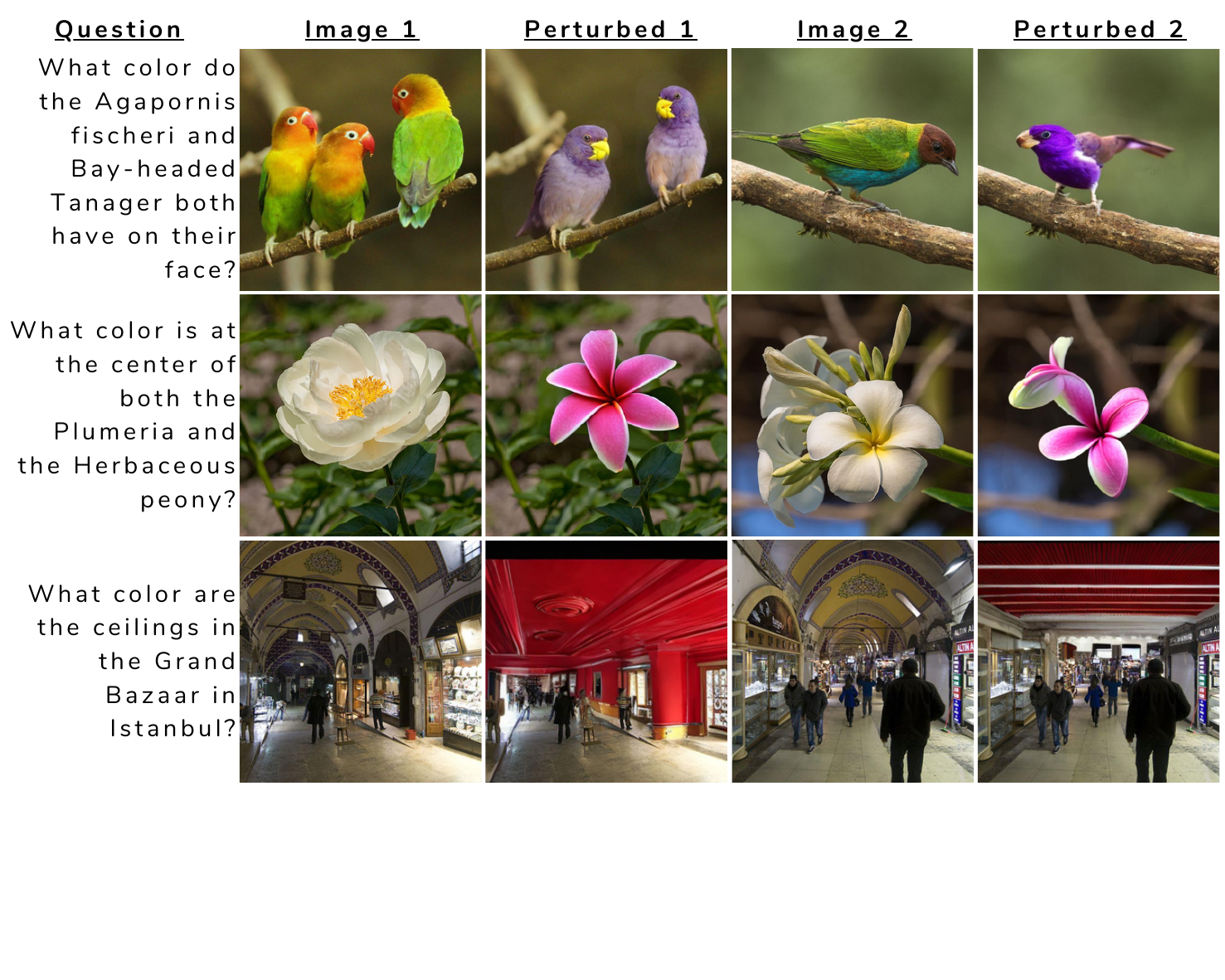}
    
    \caption{Examples of original and perturbed images in the \segsub validation set\update{, demonstrating parametric and source conflicts}. Baseline samples are comprised of image 1 and 2. Perturbed examples are comprised of perturbed image 1 and 2. Conflicting samples are comprised of (image 1, perturbed image 2) and (perturbed image 1, image 2).}
    \label{fig:perturbed_examples}
\end{figure*}

We look at three main types of conflicts between different sources of information, and study the reasoning abilities of different models on them. Note, we adopt the concept of the retrieval token \retlabel from Labruna et. al.\citep{labruna2024retrieve}.

\noindent(i) \textit{Counterfactual conflicts}: We invalidate the premise of the question by removing the object in question from the image source. As a result, any answer except for requests for more information, or statements about lacking information (\retlabel) are incorrect (\autoref{fig:counterfactual_examples}).

\noindent(ii) \textit{Parametric conflicts}: We introduce conflicts between the encoded knowledge (embedded in the learned weights) and the input image source. To study this effect, we alter attributes like the shape or color of the object under consideration in the image, therefore changing the expected response to the new label, \updatedlabel. This requires the model to rely on the new image and ignore any learned knowledge it may have about the image to answer the question correctly (for example, \autoref{fig:perturbed_examples}).
    
\noindent(iii) \textit{Source conflicts}: We introduce contradictions between the input images such that the question becomes unanswerable. For multihop questions (i.e. questions with two image sources), we augment that dataset by combining the perturbed variant of one of the two images with the original version of the other i.e. (image 1, perturbed image 2) and vice versa. By making the question unanswerable, the retrieval token \retlabel is the only correct response (see \autoref{fig:perturbed_examples}).

Each conflict type is grounded in specific hallucination mechanisms defined by prior work \citep{huang_survey_2025}: parametric conflicts manifest as factual contradictions where models prioritize encoded parametric knowledge over visual evidence; source conflicts reveal logical inconsistency hallucinations when reasoning across multiple, conflicting information sources; and counterfactual conflicts trigger extrinsic hallucinations, where models hallucinate details that can't be verified because of missing context in images. While our primary focus is on model hallucination patterns, these conflict types also relate to frameworks of information disorder \citep{wardle2017information,tandoc_defining_2018,fallis2015disinformation} which categorize various types of factual inconsistencies.

\subsection{The Knowledge Conflicts Dataset}
Existing VQA datasets lack samples with knowledge conflicts. To address this, we take three popular VQA datasets, WebQA \citep{chang_webqa_2021}, VQAv2 \citep{goyal2017making}, and OK-VQA \citep{marino_ok-vqa_2019} (see \autoref{tab:org_dataset}), and augment them with knowledge conflicts by perturbing the image sources and updating the expected answers using the \segsub framework.

We generate counterfactual samples for all three datasets. Unlike WebQA, where questions fall into specific categories (color, shape, yes-no, number), VQAv2 on OK-VQA are open-domain tasks. As a result, we can only use feature modifications to generate parametric conflicts for the WebQA dataset (as in \autoref{fig:seg_sub_pipeline}, \autoref{fig:perturbed_examples}). In addition, we can only generate source conflicts for the multihop (two-image) WebQA samples. 
% We cannot generate source conflicts for VQAv2 and OK-VQA as they are single-image VQA tasks. 

\autoref{tab:dataset_description} gives a breakdown of the samples generated for each dataset along with the method used. Note that for every perturbed sample, we include the original, unperturbed sample from the constituent dataset. This ensures that models finetuned on the generated knowledge conflicts dataset do not regress on their ability to answer traditional VQA style questions. As such, only 38\% of the resulting generations have the answer \retlabel.

% perturbations that modify object attributes such as color and shape would not result in a new label. As a result, we focus on generating counterfactual samples on VQAv2 and OK-VQA. 

\begin{table*}[t]
    \caption{A breakdown of the generated knowledge conflicts dataset by the constituent datasets, the total number of generations, and the number of generations that pass the quality checks along with label quality rating from manual evaluation.}
    \label{tab:dataset_description}
    \centering
    \resizebox{\textwidth}{!}{\begin{tabular}{llllrrr}
    \toprule
        
        \multirow{2}{*}{\textbf{Dataset}} & \multirow{2}{*}{\textbf{Conflict Type}} & \multirow{2}{*}{\textbf{Method}} & \multirow{2}{*}{\parbox{1.5 cm}{\textbf{New \\ Answer}}} & \multicolumn{2}{r}{\textbf{\# Generations: train (validation)}} & \multirow{2}{*}{\parbox{2.3 cm}{\textbf{Label Quality\\ Rating}}}  \\\cline{5-6}
        & & & & \textbf{Pre Quality} & \textbf{Post Quality} &  \\
        \midrule
        WebQA(Color, Shape) & Parametric & object infill & \updatedlabel & 141003 & 12537~(1459) & 76\% \\ 
        WebQA(Color, Shape) & Source & object infill & \retlabel & 141003 & 8038~(1050) & 82\% \\ 
        WebQA(Yes/No) & Counterfactual & object removal & \retlabel & 11077 & 1815~~~(257) & 87\% \\  
        VQAv2 & Counterfactual & object removal & \retlabel & 49742 & 7765~(1830) & 92\% \\
        OK-VQA & Counterfactual & object removal & \retlabel & 4648 & 0~~~(474) & 93\% \\
        \midrule
        Total Generations & -- & -- & -- & 201822 & 30155 (5070) & --\\
        %\midrule
        %WebQA & None & None & -- & -- & 8634 (1081) & -- \\
        %VQAv2 & None & None & -- & -- & 7765 (1830) & -- \\
        %OKVQA & None & None & -- & -- & 0 (474) & -- \\
        %\midrule
        %Total Dataset & -- & -- & -- & -- & 46554 (8455) & -- \\
        \bottomrule
    \end{tabular}}
\end{table*}

\paragraph{Quality Checks}
The generative methods used for perturbing images are imperfect. We therefore apply quality checks to filter out the noisy generations before finetuning VQA models. We present each generated sample to a quantized Qwen2-VL-7b-Instruct VLM and ask whether the modified feature is the same (or for object removal, whether the object exists), in both the original and perturbed images.
Framing the question in this way eliminates bias towards affirmative responses. Manual evaluation of the quality-checked images finds that they are indeed high quality (\autoref{tab:dataset_description}). Quality checks prompts are listed in the supplementary (\autoref{frame:quality_check_prompt_webqa_color_shape}).

\subsection{Finetuning on \segsub Data}
To evaluate the \segsub frameworks efficacy in developing VLM robustness, we finetune three VLMs on the generated knowledge conflicts data---Llava-1.5-7b \citep{liu2024improved}, Phi3-vision-128k-instruct \citep{abdin2024phi}, and Qwen2-VL-7B-Instruct \citep{wang2024qwen2}. All models are finetuned on the training set (\autoref{tab:dataset_description}) for 1 epoch on 2x NVIDIA RTX A6000 GPUs using SWIFT \cite{zhao2024swiftascalablelightweightinfrastructure}, with convergence shown in the appendix (\autoref{fig:training_loss}). We apply LoRA \citep{hu2021lora} to reduce GPU memory requirements and use Distributed Data Parallel methods DeepSpeed \citep{rasley2020deepspeed} and ZeRO \citep{rajbhandari2020zero} to train across multiple GPUs. Refer to \autoref{tab:hyperparameters} for hyperparameters.

\subsection{Evaluation}
We compare performance of the finetuned versions of the VLMs against their base versions on the \segsub validation set (\autoref{tab:dataset_description}). We also evaluate on---Llava-1.5-13b \citep{liu2024improved} and GPT-4o-mini \citep{achiam2023gpt}. \update{For each of the questions in the dataset, we  explicitly instruct the LLM to answer the question based on the provided images which aims to reduce the bias of LLMs to answer from their parametric knowledge.}

\paragraph{Evaluation on \segsub Generations}
We measure the VLM's reasoning ability over conflicting sources of information with the following accuracy scores (see the appendix for details on accuracy metrics).

\textit{Parametric response rate}: \% of model responses that incorrectly predict the original label when a color or shape attribute has been changed. Therefore, highlighting the effect of parametric conflicts on model performance by showcasing the model's over reliance on the encoded parametric knowledge instead of adapting to the modified image source.

\textit{Accuracy for counterfactual conflicts}: \% of model responses that correctly generate \retlabel~~or any response which acknowledges the models failure to answer on the set of counterfactual samples\footnote{We consider VLM responses that make a reference to not having enough information or context, being unable to make a determination, or the image source being obscured in some way as `acknowledgement' responses, equivalent to \retlabel (i.e. \autoref{tab:ret_acknowledged} in the appendix).}.

\textit{Accuracy for source conflicts}: \% of model responses that correctly generate \retlabel or any response which acknowledges the models failure to answer on the set of source conflicts.  See \autoref{tab:ret_acknowledged} in the appendix for the `acknowledgment' phrases parsed from model responses.

% a blank, all-white image. 
% which have been fine-tuned on the datasets we're looking at?

\paragraph{Evaluation on Original Samples}
We evaluate model accuracy on original samples to check for performance regressions on the original VQAv2, OK-VQA, and WebQA validation sets that may occur as a result of finetuning. Accuracy scores on the original samples are simply the \% of model responses that generate the original labels in each dataset when presented with the original, unperturbed images. These results are reported alongside accuracy scores for the knowledge conflict tasks. 
%and That is, for each generated knowledge conflict, we include the original unperturbed sample.
% \todo{WHAT? Note, for each combination of perturbation and dataset, we compare set of original samples that correspond to the set of perturbed samples.}

\paragraph{Robustness on Counterfactuals}
Counterfactual conflicts are generated using LaMa. To ensure that our finetuned models do not learn to predict \retlabel~ based on whether or not the image was modified by LaMa, we include an additional robustness check. For each perturbed counterfactual image and question pair in the WebQA dataset, we create randomized counterfactual samples by pairing a question with an unaltered image sampled at random from the WebQA dataset. We call these randomized, negatively sampled counterfactuals.

\paragraph{Image-Question Contextualization}
Finally, we analyze the effect of contextualization on hallucination rate. As such, we prompt GPT-4o-mini to assign a `contextualization score' to each counterfactual image and question pair in the \segsub validation set (see \autoref{frame:image_question_contextualization} in the supplementary). Intuitively, this concept relates to the amount of contextual cues that an image has for a given question, i.e. the more the number of contextual cues an image has, the more hints the model has to answer the given question. For highly contextualized image question pairs, visual reasoning is reinforced by various elements within the image that prime the model to hallucinate. In poorly contextualized pairs, image sources lack the context cues that exhibit this priming effect, and therefore do not provoke hallucinations.

\section{Dataset Quality Checks} % 1.5 - 3
After generating a large number of samples (\textgreater$200,000$), we apply quality checks to remove noisy generations, resulting in approximately $35,225$ samples. Two raters independently labeled a subset of $100$ quality-checked generations for each category of conflicts to determine if the new label (\retlabel or \updatedlabel) matches the perturbed image---see label quality ratings in \autoref{tab:dataset_description}.

Counterfactuals have a higher quality rating (\textgreater90\%). Parametric (76\%) and source conflicts (82\%) produce more noisy generations which we attribute to the increased difficulty in replacing an object versus removing it. Raters only disagreed on a small fraction of samples (30/300), while a Cohen's Kappa of 0.45 reflects that disagreements happened only on lower quality generations \cite{delgado2019cohen}. See \autoref{fig:counterfactual_examples} for examples of counterfactual image generations and \autoref{fig:perturbed_examples} for parametric and source conflicts.

% However, since these disagreements are on the  ~\cite{delgado2019cohen} this results in a Cohen's Kappa of 0.45. 

%The low Kappa score is a consequence of class imbalance ~\cite{delgado2019cohen} which, for instance, can be seen from the high quality rating score (87-93\%) for counterfactual generations.

% \begin{figure*}
%     \centering
%     \vspace{-4mm}
%     \includegraphics[width=0.8\linewidth]{figures/results/finetuning_eval.pdf}
%     \vspace{-2mm}
%     \caption{Evaluation of baseline (-Base) and \segsub finetuned (-Ft) model accuracy on counterfactual and source conflicts (higher is better). Evaluation on original samples from VQAv2, OK-VQA, and WebQA datasets shows that finetuning does not result in performance regression on these tasks (except on WebQA two-image samples). Finetuned models outperform baselines across all types of knowledge conflict.}
%     \vspace{-4mm}
%     \label{fig:finetuning_results}
    
% \end{figure*}

\section{Results} % 1.5 - 3

\begin{figure*}
    \centering
    \begin{minipage}[t]{0.48\textwidth}
        \centering
        \includegraphics[width=0.9\linewidth]{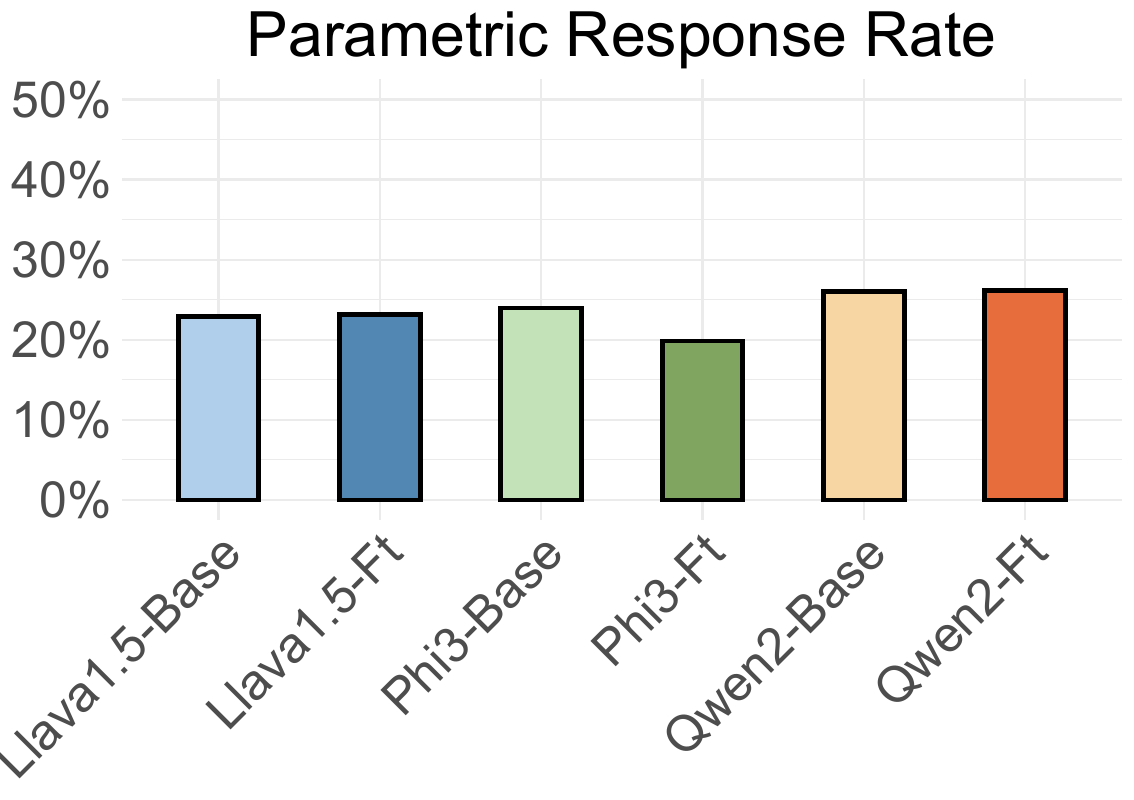}
        \vspace{-4mm}
        \caption{Parametric effect analysis: how often does the model predict the original label for perturbed images? Lower is better, implying a reduced parametric effect.}
        \label{fig:parametric_effect}
    \end{minipage}%
    \hfill
    \begin{minipage}[t]{0.48\textwidth}
        \centering
        \includegraphics[width=\linewidth]{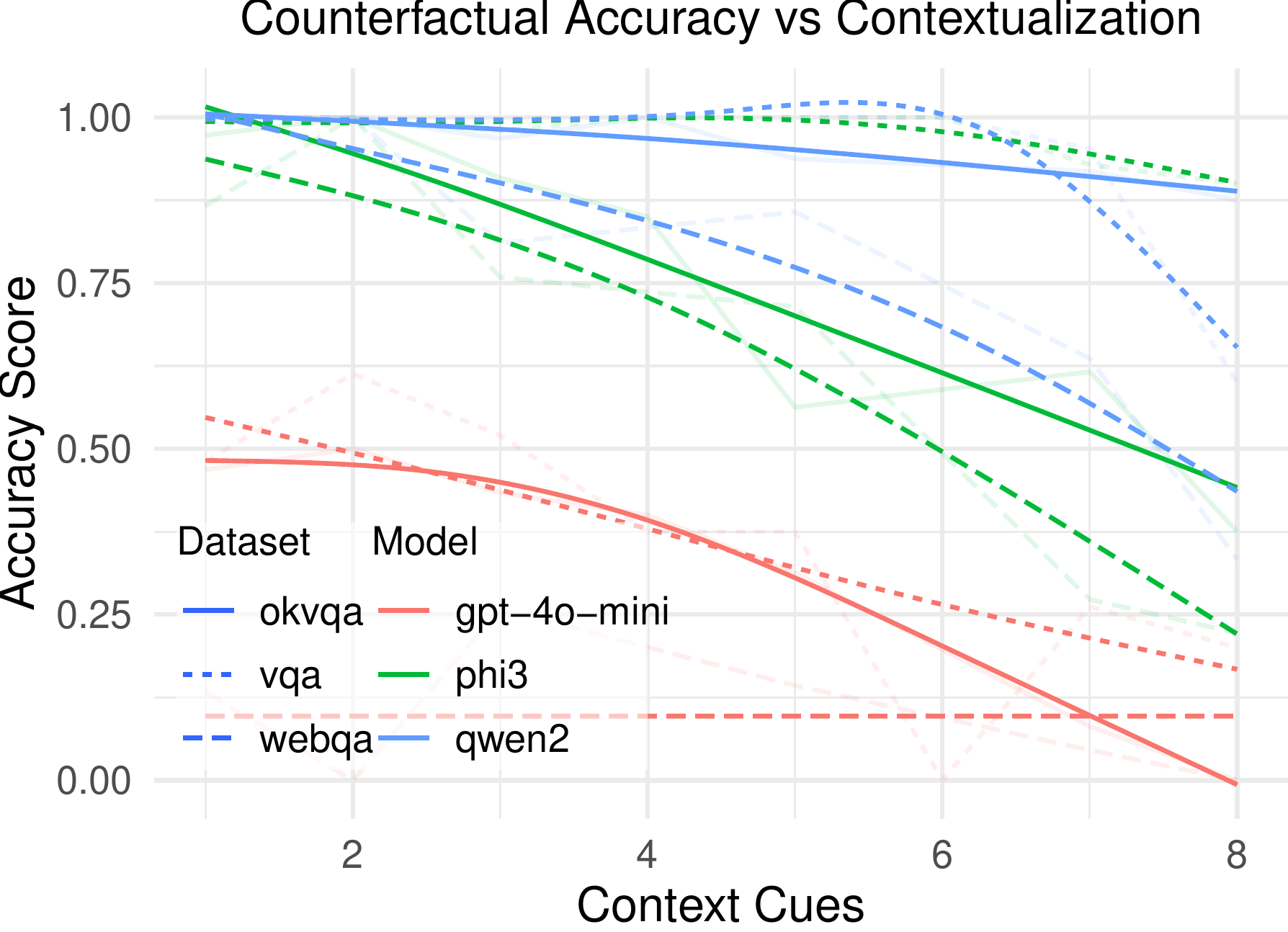}
        \caption{Decreased Accuracy on Counterfactual Conflicts in finetuned VLMs (and GPT4-o-mini) with Increasing Image Contextualization Scores. Baseline unsmoothed data is in the background.}
        \label{fig:context_scores}
    \end{minipage}
\end{figure*}

\begin{figure*}
    \centering
    \vspace{-4mm}
    \includegraphics[width=0.8\linewidth]{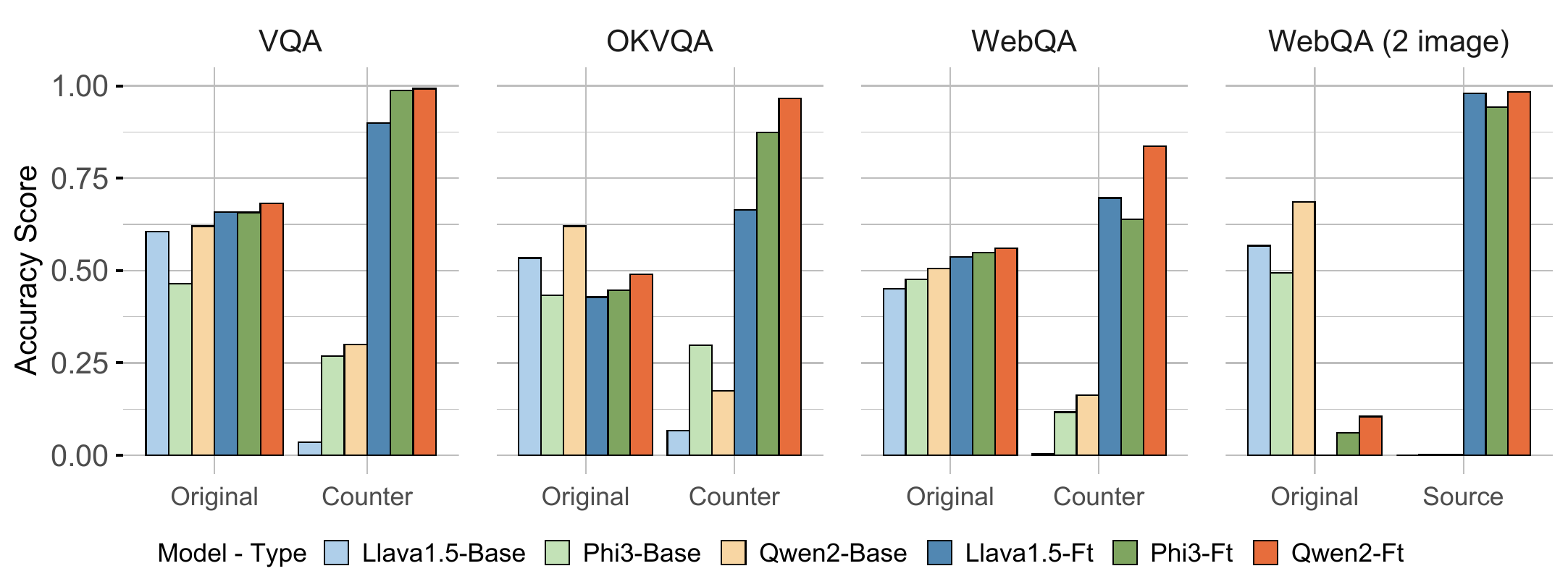}
    \vspace{-2mm}
    \caption{Evaluation of baseline (-Base) and \segsub finetuned (-Ft) model accuracy on counterfactual and source conflicts (higher is better). Evaluation on original, un-perturbed samples from VQAv2, OK-VQA, and WebQA datasets ("Original") shows that finetuning does not result in performance regression on these tasks (except on WebQA two-image samples). Finetuned models outperform baselines across counterfactual ("Counter") and source conflicts.}
    \vspace{-4mm}
    \label{fig:finetuning_results}
\end{figure*}

In \autoref{fig:finetuning_results} we find that baseline VLMs fail to acknowledge counterfactual conflicts (Counter) and source conflicts (Source). Finetuning mitigates this across every dataset. The resulting finetuned models (-Ft) outperform the baseline models (-Base) on perturbed samples. Finetuning has some benefit on the original samples (Original) for VQA and WebQA counterfactual sources, but a large performance regression is apparent for samples with source conflicts in WebQA. Models are generally robust to parametric conflicts.

\vspace{-2mm}
\paragraph{Parametric Conflicts}
While Phi3 model does benefit somewhat from finetuning (4\% drop in parametric response rate), parametric response rates are low for both baseline and finetuned models ($\sim$20\%, \autoref{fig:parametric_effect}). As such, baseline models are already robust to conflicts between image sources and parametric memory.

% \begin{figure}
%     \centering
   
%     \includegraphics[width=0.8\linewidth]{figures/results/parametric_plot.pdf}
%     % \includegraphics[width=0.9\linewidth]{figures/results/perturbed_1_vs_2_image.pdf}
%     \vspace{-4mm}
%     \caption{Parametric effect analysis: how often does the model predict the original label for perturbed images? Lower is better, implying a reduced parametric effect.}
%     \label{fig:parametric_effect}
%     \vspace{-2mm}
% \end{figure}

% \begin{figure}
% \centering
%     \includegraphics[width=0.9\linewidth]{figures/results/context_scores.pdf}
%     % \vspace{-3mm}
%     \caption{Decreased Accuracy on Counterfactual Conflicts in finetuned VLMs (and GPT4-o-mini) with Increasing Image Contextualization Scores. Baseline unsmoothed data is in the background.}
%     \label{fig:context_scores}
%     % \vspace{-4mm}
% \end{figure}

\vspace{-2mm}
\paragraph{Source Conflicts}
For WebQA samples with source conflicts, the finetuned models have extremely low accuracy on original samples. This is a result of the finetuned models failing to predict the old label and instead overpredicting the \retlabel when presented with two images. %We attribute this to the high amount of noise with knowledge conflict generations achieving a quality rating of only 82\% (\autoref{tab:dataset_description}), and the multihop nature of the knowledge conflict task. 
Interestingly, instead of generating an `acknowledgement' response, baseline models tend to predict one of the two incorrect answers---either the original label (for the unperturbed image) or \updatedlabel (for the perturbed image)---uniformly at random. 
% See \autoref{tab:webqa_conflicts} in the supplementary.

\vspace{-2mm}

\paragraph{Counterfactual Conflicts}
Baseline models perform poorly on counterfactual conflicts, with no model achieving more than 30\% accuracy. Since these models are not trained to return the \retlabel, we consider any admission of failure by the model as a \retlabel. While baseline models are sometimes able to determine when an image lacks the information required to answer a question, they are not robust to these samples. Our key insight is that finetuning enables these models to identify counterfactual conflicts with high accuracy, without degrading performance on the original datasets. Additionally, finetuning provides a 5-10\% performance gain on the original samples from WebQA and VQA datasets.

\vspace{-2mm}
\paragraph{Robustness of Counterfactual Conflicts}
We find that finetuned models are robust in detecting randomized counterfactual samples and are not simply detecting images that have been modified using \segsub. The finetuned Qwen2 model predicts \retlabel for 80\% of the randomized counterfactuals sampled from the WebQA dataset. \autoref{tab:natural_counterfactual_results} in the appendix has further details.

\vspace{-2mm}

\paragraph{Parameter Size}
We find that performance improvements on the evaluation metrics derived from increasing model size have diminishing returns. Lightweight finetuned models still outperform SoTA models such as GPT-4o (see \autoref{fig:baseline_model_performance} in the appendix).

\begin{figure}
    \centering
    \begin{subfigure}[b]{0.22\textwidth}
        \centering
        \includegraphics[width=\textwidth]{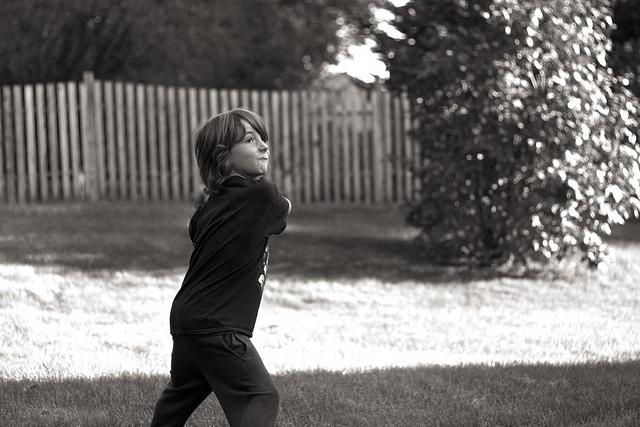}  % Replace with your figure file
        \caption{ChatGPT: "There doesn’t appear to be an object clearly visible in his hands."}
        \label{fig:bat_boy}
    \end{subfigure}%
    \hfill
    \begin{subfigure}[b]{0.22\textwidth}
        \centering
        \includegraphics[width=\textwidth]{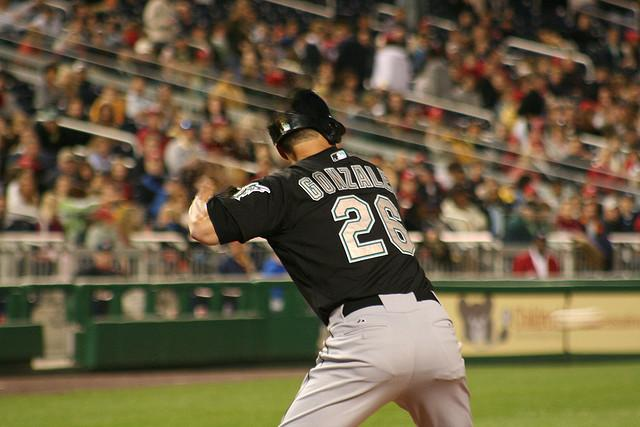}  % Replace with your figure file
        \caption{ChatGPT: "The batsman in the image is holding a baseball bat as he prepares to swing."}
        \label{fig:bat_batsman}
    \end{subfigure}
    \caption{These counterfactual examples were generated by removing a baseball bat from two different VQA images. When asked `what is he holding?', ChatGPT only hallucinates in the highly contextualized case (right).}
    \label{fig:baseball_example}
\end{figure}

\paragraph{Image-Question Contextualization}
Contextual cues within an image provide models with clues to answer the question, as in \autoref{fig:baseball_example}. We find statistically significant negative correlation between image-question contextualization, as approximated by GPT-4o-mini, and accuracy on counterfactual samples ($r = -0.368, p = 0.003$). \autoref{fig:context_scores} reveals that models perform poorly in identifying a sample as counterfactual (i.e. lower accuracy of predicting \retlabel) and is more likely to hallucinate on heavily contextualized image question pairs. Interestingly, GPT-4o-mini hallucinates for all of the counterfactual examples given in \autoref{fig:counterfactual_examples}.

For a concrete example, see \autoref{fig:baseball_example}, where both counterfactual examples were generated by removing a baseball bat. Here, a poorly contextualized image question pair features a child standing in a field with the question "what is he holding?" (\autoref{fig:bat_boy}). The only contextual cues as to what the child might have been holding are the generic outdoor setting, and the child's pose. Contrasting this in the adjoining sample is a baseball player, adorned in a jersey with his player number printed on the back, in a stadium filled with baseball fans (\autoref{fig:bat_batsman}). ChatGPT recognizes that the child is holding nothing but hallucinates a bat in the hands of the batsman. Alongside previous works that show a relationship between image context and object detection \citep{beery2018recognition}, these results indicate that contextual cues have a priming effect that induces hallucinations in VLMs for highly contextualized counterfactuals.

\section{Discussion}

The \segsub framework extends research on reasoning with knowledge conflicts to the multimodal domain. The framework builds on the unimodal text-based Entity Replacement Framework \citep{longpre_entity-based_2022} and extends it to VQA by segmenting and modifying relevant entities and objects in images. Our perturbations are inspired from prior work on knowledge conflicts \citep{chen_rich_2022,longpre_entity-based_2022},  counterfactual reasoning \citep{neeman_disentqa_2022,hong_why_2024} and hallucinations~\citep{huang_survey_2025} in LLMs. 

VLMs, like LLMs, may internalize statistical and factual knowledge from large-scale training data. This includes details such as the typical colors of specific bird and flower species (\autoref{fig:perturbed_examples}) or even historical facts such as the color of the horse that Eli Bremer rode in the 2008 Summer Olympics (\autoref{fig:seg_sub_pipeline}). We measure the degree to which VQA models prioritize these parametric facts over the information contained in input sources. Contrary to prior findings, where LLMs are more susceptible to parametric responses when presented with knowledge conflicts \citep{hong_why_2024}, we find that VLMs are less prone to parametric responses with $\sim$20\% across all models tested (\autoref{fig:parametric_effect}).

Our framework is also motivated by an over-reliance on paired image-caption data and contrastive loss functions for training VLMs. Image-caption datasets help models learn to reason about what is in the image, for example, in image summarization tasks. Perhaps because of this over-reliance, we find that models struggle with reasoning about what \textit{is not} in the image. While some of the examples in \autoref{fig:counterfactual_examples} seem unusual at first glance, they highlight a fundamental gap that exists in how models and humans perceive these tasks. Without finetuning, models such as GPT-4o ignore the counterfactual sources and often hallucinate (\autoref{fig:context_scores}). Our work aims to address this gap by introducing counterfactual samples into the training process.  

We find that counterfactual reasoning in VLMs is conditional on the sources presented. When presented with `randomized negatively sampled counterfactuals' both the base and finetuned models can correctly determine that the images are unrelated to the query (\autoref{tab:natural_counterfactual_results}). In contrast, sources with many contextual cues cause models to hallucinate. This finding is consistent across all models and VQA datasets tested, and it aligns with what \citet{huang_survey_2025} identify as "over-confidence" hallucinations. By maximizing token likelihood, models prioritize contextual coherence at the expense of factual accuracy. When visual contexts contain rich semantic cues, models are primed to hallucinate the missing content rather than acknowledge information gaps. This link between over-confidence hallucinations and highly contextualized counterfactual samples underlines the value of our framework and dataset for understanding and mitigating hallucination mechanisms in multimodal reasoning.

\section{Ethical Considerations}
% Parametric misinfo

At its most fundamental level, misinformation arises from knowledge conflicts between factual knowledge and non-factual information. While on the surface \segsub may seem directly applicable in this domain, generated samples in the \segsub dataset fall under the category of false information and not misinformation. In general, there is a gap between the domains covered by VQA datasets and topics that cause societal harms. Such harms stem from the likelihood of a piece of false information changing existing beliefs or forming new beliefs \cite{fallis2015disinformation}. While knowledge conflicts may represent necessary preconditions for misinformation, they are insufficient. Critical factors that go beyond epistemology, such as creator intent, potential harm, and contextual appropriateness, must be considered. These necessitate the study of propagation dynamics, audience susceptibility, or content amplification mechanisms.

Despite this, many of our findings are relevant to this field. First, concerns over the AI-generated misinformation stem from training on factually incorrect information and later regurgitating it through learned responses \citep{huang_survey_2025}. Low parametric response rates should ease concerns for typical VQA tasks (\autoref{fig:parametric_effect}). Second, we find that current models are unable to identify the existence of knowledge conflicts. This is problematic regardless of which source is deemed correct, as systems that cannot flag knowledge conflicts will never be able to recover from them. Finally, the determination of a correct source among conflicting samples is domain and context-specific, and should be decided not by those training a general-purpose model, but by system designers. In practice, the \ret~ label may trigger a retrieval component to supply additional context \citep{labruna2024retrieve}, and system designers can decide whether they have a greater level of trust in the training data or the sources used at inference.

\subsection{Limitations and Future Work} % 1.5 - 3
\label{sec:limitations}
Our framework effectively generates and evaluates parametric, source, and counterfactual conflicts across VQA datasets. However, three key limitations may affect its generalizability: reliance on VLMs for quality checks, residual and generative artifacts, and image-question contextualization.

First, we rely on smaller quantized VLMs for quality assurance which may introduce an additional source of error. A fine-grained visual and semantic understanding in the VLM could lead to overlooked errors in perturbation or segmentation that affect the dataset’s overall quality. Although we manually review a subset of outputs from each perturbation type to gauge quality, the effectiveness of quality control could be enhanced by leveraging more powerful models or ensemble-based methods. We also note the possibility of the quality-check ruling out high quality generations. However, this is less of a concern as we wish to minimize false positives in the dataset, and we can compensate simply by generating more samples.

Second, handling residual artifacts left after object removal, like shadows or reflections, is challenging. These artifacts can indicate the previous presence of objects, introducing noise and inconsistencies that may mislead models that are sensitive to visual details. While we mitigate this partially through manual evaluation and quality checks, future work could explore advanced inpainting or shadow removal for cleaner counterfactuals. Current generative methods suffer from quality issues, with artifacts like blurred infilled regions and excessive noise in segmented areas, despite high quality ratings across perturbation categories. Emerging text-to-image editing models \citep{hui_hq-edit_2024,bodur_iedit_2023,zhang_magicbrush_2024} may help address these issues. 
% While we employ a rule-based segmentation approach, these models dynamically infer infill regions from input prompts. Given the lower quality ratings for knowledge conflict perturbations, future work should explore new generative methods to improve this aspect.

Finally, our evaluation of hallucinations is restricted to model responses on counterfactual samples. While the narrow framing of VQA datasets enables measurement, it ignores the causes of hallucination that emerge from training and inference procedures \citep{huang_survey_2025}. In addition, while our image-question contextualization analysis relies on GPT-4o-mini, more technical methods could be developed. For instance, one could generate question sets for each image and compute text similarity between the dataset questions and generated questions. We leave this as the subject of future work.

Without our framework, visual knowledge conflicts are difficult and costly to collect. Alternatives approaches that aim to identify counterfactual image sources instead of using a generative approach would require image retrieval systems that use fact-based reasoning rather than similarity measures. Our generative approach provides a systematic way for future work to build on counterfactual reasoning, source conflicts, and hallucinations in the multimodal setting. Future work may center around developing more sophisticated sets of generative constraints and extending the \segsub framework and dataset to tackle aspects of visual reasoning that continue to be underrepresented in VQA datasets.

\section{Conclusion} % 1.5 - 3
% In this work, we introduced \segsub, a Segmentation Substitution framework designed to improve the robustness of visual reasoning in VLMs. Through the application of image segmentation and inpainting techniques, we augment VQA datasets with counterfactual samples and knowledge conflicts. These samples test LLMs' abilities to recognize and respond to various types of image-based reasoning challenges. Our experiments demonstrate that while VLMs show resilience to certain perturbations such as feature modifications that lie within their training distribution, they struggle with counterfactual cases and inconsistencies across multiple image sources, especially in multi-hop scenarios.

% Our findings highlight the need for robust VQA models that can navigate diverse visual contexts. We hope our contribution to advancing visual reasoning and model resilience against counterfactual noise will encourage future research in this area. 

% The \segsub framework serves as a tool for strengthening VQA tasks, advancing the study of multimodal reasoning in real-world applications.

We introduce \segsub, a framework designed to improve the robustness of visual reasoning in VLMs. Through the application of image segmentation and inpainting techniques, we augment VQA datasets with parametric, source and counterfactual conflicts. These samples test VLMs' abilities to recognize and respond to various types of image-based reasoning challenges. While our experiments demonstrate VLM resilience to perturbations that lie within their training distribution (i.e. feature modifications that induce parametric conflicts), they struggle with counterfactual cases and conflicts across multiple image sources, especially in multi-hop scenarios. These findings, as well as the accompanying benchmark dataset, highlight critical areas for enhancing VLM training protocols, particularly in addressing complex knowledge conflicts that induce hallucinations in information-sensitive contexts.

% Our findings highlight the need for VQA models that are robust to knowledge conflicts and we hope that our contribution will inspire future research in advancing visual reasoning. 

% Shorter
% We introduced \segsub, a framework that enhances visual reasoning in VLMs by augmenting VQA datasets with counterfactual samples and knowledge conflicts through segmentation and inpainting. Our experiments show that while VLMs are resilient to perturbations within their training distribution, they struggle with counterfactuals and inconsistencies in multi-image contexts, particularly in multi-hop tasks. These results emphasize the need for more robust VQA models capable of handling diverse visual challenges, encouraging further research in counterfactual noise and visual reasoning

% The \segsub framework serves as a tool 

% \section{Ethical Considerations}
% \input{sections/10-ethics}

\section{Acknowledgements}
This work was supported in part by the Office of Naval Research grant (N000141812106) and the Knight Foundation. Additional support was provided by the Center for Computational Analysis of Social and Organizational Systems (CASOS) at Carnegie Mellon University. The views and conclusions contained in this document are those of the authors and should not be interpreted as representing the official policies, either expressed or implied, of the Knight Foundation, Office of Naval Research, or the U.S. Government.

\bibliography{bibs/WebQA,bibs/Retrieval,bibs/Reasoning,bibs/other,bibs/entity_replacement}
% \bibliographystyle{acl_natbib}

% \break
\section{Ethics Checklist}
\begin{enumerate}

\item For most authors...
\begin{enumerate}
    \item  Would answering this research question advance science without violating social contracts, such as violating privacy norms, perpetuating unfair profiling, exacerbating the socio-economic divide, or implying disrespect to societies or cultures? Yes
  \item Do your main claims in the abstract and introduction accurately reflect the paper's contributions and scope? Yes
   \item Do you clarify how the proposed methodological approach is appropriate for the claims made? Yes
   \item Do you clarify what are possible artifacts in the data used, given population-specific distributions? Yes
  \item Did you describe the limitations of your work? Yes
  \item Did you discuss any potential negative societal impacts of your work? Yes
      \item Did you discuss any potential misuse of your work? Yes
    \item Did you describe steps taken to prevent or mitigate potential negative outcomes of the research, such as data and model documentation, data anonymization, responsible release, access control, and the reproducibility of findings? Yes
  \item Have you read the ethics review guidelines and ensured that your paper conforms to them? Yes
\end{enumerate}

\item Additionally, if your study involves hypotheses testing... N/A
\begin{enumerate}
  \item Did you clearly state the assumptions underlying all theoretical results?
    N/A
  \item Have you provided justifications for all theoretical results?
    N/A
  \item Did you discuss competing hypotheses or theories that might challenge or complement your theoretical results?
    N/A
  \item Have you considered alternative mechanisms or explanations that might account for the same outcomes observed in your study?
    N/A
  \item Did you address potential biases or limitations in your theoretical framework?
    N/A
  \item Have you related your theoretical results to the existing literature in social science?
    N/A
  \item Did you discuss the implications of your theoretical results for policy, practice, or further research in the social science domain?
    N/A
\end{enumerate}

\item Additionally, if you are including theoretical proofs... N/A
\begin{enumerate}
  \item Did you state the full set of assumptions of all theoretical results?
    N/A
	\item Did you include complete proofs of all theoretical results?
    N/A
\end{enumerate}

\item Additionally, if you ran machine learning experiments...
  \begin{enumerate}
  \item Did you include the code, data, and instructions needed to reproduce the main experimental results (either in the supplemental material or as a URL)?
    Yes
  \item Did you specify all the training details (e.g., data splits, hyperparameters, how they were chosen)?
    Yes
     \item Did you report error bars (e.g., with respect to the random seed after running experiments multiple times)?
    No
	\item Did you include the total amount of compute and the type of resources used (e.g., type of GPUs, internal cluster, or cloud provider)?
    Yes
     \item Do you justify how the proposed evaluation is sufficient and appropriate to the claims made? 
    Yes
     \item Do you discuss what is ``the cost`` of misclassification and fault (in)tolerance?
    Yes
  
\end{enumerate}

\item Additionally, if you are using existing assets (e.g., code, data, models) or curating/releasing new assets, \textbf{without compromising anonymity}...
\begin{enumerate}
  \item If your work uses existing assets, did you cite the creators? Yes
  \item Did you mention the license of the assets? Yes
  \item Did you include any new assets in the supplemental material or as a URL? Yes
  \item Did you discuss whether and how consent was obtained from people whose data you're using/curating? N/A
  \item Did you discuss whether the data you are using/curating contains personally identifiable information or offensive content? N/A
\item If you are curating or releasing new datasets, did you discuss how you intend to make your datasets FAIR? N/A
\item If you are curating or releasing new datasets, did you create a Datasheet for the Dataset? Yes
\end{enumerate}

\item Additionally, if you used crowdsourcing or conducted research with human subjects, \textbf{without compromising anonymity}... N/A
\begin{enumerate}
  \item Did you include the full text of instructions given to participants and screenshots?
    N/A
  \item Did you describe any potential participant risks, with mentions of Institutional Review Board (IRB) approvals?
    N/A
  \item Did you include the estimated hourly wage paid to participants and the total amount spent on participant compensation?
    N/A
   \item Did you discuss how data is stored, shared, and deidentified?
   N/A
\end{enumerate}

\end{enumerate}
% \clearpage
% \pagebreak

\appendix

\section{Appendix}

% \begin{figure}
%     \centering
%     \includegraphics[width=0.5\linewidth]{figures/results/negative_eval.pdf}
%     \caption{Full results for the randomized negatively sampled robustness check. Models finetuned on \segsub data (-ft) outperform baseline models in identifying images irrelevant to the given query.}
%     \label{fig:negatively_sampled_counterfactual_eval}
% \end{figure}

\subsection{Model Finetuning}

\begin{figure}
    \centering
    \begin{subfigure}{0.4\textwidth}
        \includegraphics[width=0.98\linewidth]{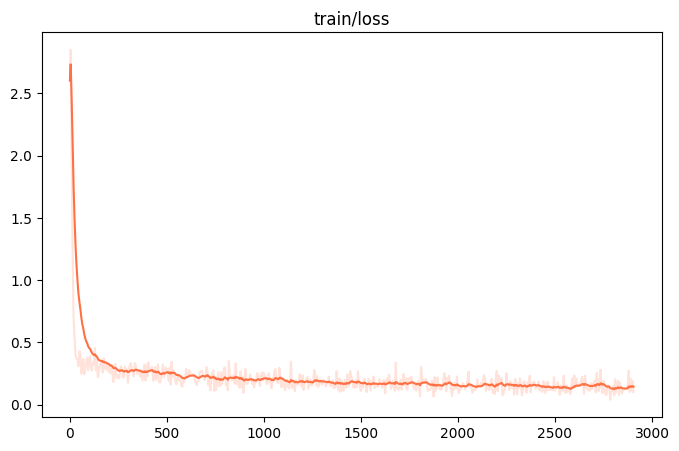}
    \end{subfigure}
    \begin{subfigure}{0.4\textwidth}
        \includegraphics[width=0.98\linewidth]{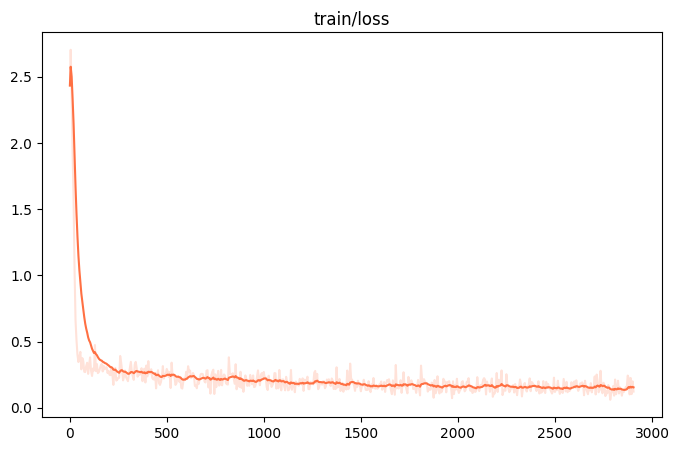}
    \end{subfigure}    
    \caption{Top: Qwen2 training loss. Bottom: Phi3 training loss.}
    \label{fig:training_loss}
\end{figure}

Hyperparameters for the finetuned models are given in \autoref{tab:hyperparameters}. Note: Clip-vit refers to openai/clip-vit-large-patch14-336. Convergence of training loss within one epoch for Qwen2 and Phi3 is shown in \autoref{fig:training_loss}.

\begin{table}[!ht]
\caption{Important hyperparameters for the models.}
\label{tab:hyperparameters}
\centering
\begin{tabular}{lccc}
\toprule
Hyperparameter & Phi3V & Qwen2VL & Llava \\
\midrule
hidden size & 3072 & 3584 & 4096 \\
hidden act & silu & silu & gelu \\
intermediate size & 8192 & 18944 & 4096 \\
\# attention heads & 32 & 28 & 16 \\
\# hidden layers & 32 & 28 & 24 \\
vision model & clip & qwen2 & clip \\
$|$image embedding$|$ & 1024 & N/A & 768 \\
vocab size & 32k & 152k & 32k \\
$|$pos. embedding$|$ & 131k & 32k & 4096 \\
torch dtype & bf16 & bf16 & f16 \\
initializer range & 0.02 & 0.02 & N/A \\
sliding window & 131k & 32k & N/A \\
temperature & 0.01 & 0.01 & 0.01 \\
\bottomrule
\end{tabular}
\end{table}

\subsection{Prompts}
Prompts for QA checks and image-question context evaluation are listed here---namely the counterfactual QA check, the feature modification QA check, and the image-question contextualization prompt.

\begin{framed}
\label{frame:quality_check_prompt_webqa_yesno}
\textbf{human}: 

$\langle$image-placeholder$\rangle$

Caption: $\langle$Original Image$\rangle$

$\langle$image-placeholder$\rangle$

Caption: $\langle$Perturbed Image$\rangle$

Question (for object removal): is the $\langle$object$\rangle$ present in both the original image and the perturbed image?

Question (for color and shape change): what is the $\langle$category$\rangle$ of the $\langle$object$\rangle$ in the image?

\textbf{ai}: 
\end{framed}

\begin{framed}
% /Inpaint-Anything/tasks/webqa/gpt_qa_check.py
\label{frame:quality_check_prompt_webqa_color_shape}
\textbf{system}: You must use the provided image sources to answer the question. If the answer is not in the image, respond 'unknown'.

\textbf{human}: 

Image: $\langle$image-placeholder$\rangle$

Caption: $\langle$caption$\rangle$

Question: $\langle$query$\rangle$

\textbf{ai}: 
\end{framed}

\begin{framed}
\label{frame:image_question_contextualization}
\textbf{system}: Give a contextualization score for each image question pair. The score, between 1 and 10, should reflect the degree to which the image contextualizes the question. That is, how likely is it that you might come up with the question while looking at the image. Focus on the range of possible questions that might be asked about the image; that is, how likely is the given question, in this entire set. Give just the score, no explanation.

\textbf{human}: 

$\langle$counterfactual-image$\rangle$

Question: $\langle$question$\rangle$

\textbf{ai}: 
\end{framed}

\subsection{Larger VLMs}
\begin{figure*}
    \centering
    \includegraphics[width=\linewidth]{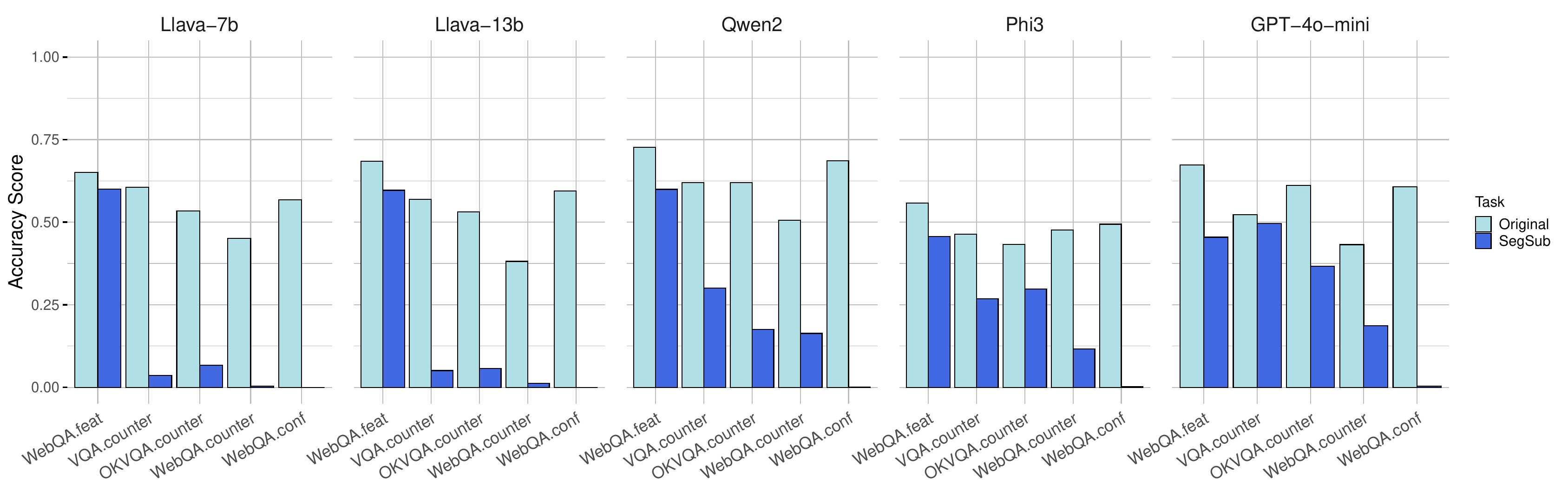}
    \caption{Baseline model performance on original and perturbed labels for the various datasets and tasks.}
    \label{fig:baseline_model_performance}
\end{figure*}

Finally, we include the accuracy of two additional baseline models, Llava-1.5-13b and GPT-4o-mini, on both the original VQA tasks and the various tasks in the \segsub dataset (\autoref{fig:baseline_model_performance}). As previously discussed, performance improvements from larger baseline VLMs are limited (Llava-7b vs Llava-13b). None of the baseline models are capable of matching the performance of \segsub finetuned models.

\subsection{Robustness Checks}
As models are not trained on irrelevant images, randomly sampling negative image query pairs from across our three datasets is an out-of-distribution task. This evaluates the robustness of our finetuning process on the more trivial cases where the image and query are irrelevant. \autoref{tab:natural_counterfactual_results} shows the full set of results, which as previously discussed reveal that finetuned models have improved performance compared with baseline models. The list of `acknowledgment' terms we consider as admissions of failure to reason over an image query pair due to incomplete information are given in \autoref{tab:ret_acknowledged}. 

Accuracy on OK-VQA negatively sampled counterfactuals is lower, which we attribute to the fact that the task itself is designed in such a way as to require knowledge external to the sources presented to the model. Future work on incorporating retrieval systems that are robust to counterfactual noise is warranted, particularly for open-domain, outside-knowledge tasks such as OK-VQA.

\begin{table}
    \caption{Full results for the randomized negatively sampled robustness check. Models finetuned on \segsub data (-ft) outperform baseline models in identifying images irrelevant to the given query.}
    \label{tab:natural_counterfactual_results}
    \centering
    \begin{tabular}{@{}llll@{}}
\toprule
Model & WebQA              & VQA                 & OK-VQA               \\ \midrule
qwen2-ft   & 0.80  & 0.62 & 0.28  \\
qwen2      & 0.11  & 0.28 & 0.07  \\
phi3-ft    & 0.36  & 0.60 & 0.27  \\
phi3       & 0.30  & 0.34 & 0.37  \\
llava15-ft & 0.24  & 0.59 & 0.28  \\
llava15    & 0.00  & 0.07 & 0.06  \\ \bottomrule
\end{tabular}
\end{table}

\begin{table}
    \caption{A list of terms that baseline models may use to express a failure to answer the given question based on insufficient information.}
    \label{tab:ret_acknowledged}
    \centering
    \begin{tabular}{l}
    $\langle$RET$\rangle$ (i.e. \retlabel) \\
    Sorry \\
    I cannot \\
    I do not \\
    image does not \\
    information \\
    not enough \\
    not clear \\
    not visible \\
    not sure \\
    not able \\
    determine \\
    blurry   \\
    blurred   \\
    no existence \\
    context \\
    apologize \\
    % white background \\
    \end{tabular}
\end{table}

\subsection{WebQA Accuracy}
\label{sec:accuracy}
Accuracy on the WebQA task is determined by comparing a restricted bag of words (bow) vector between the expected (E) and generated (G) answers;
\begin{equation}
    \label{eq:ACC}
    \text{Acc} = \frac{1}{n}\Sigma [\frac{|\text{bow}_{E} \cap \text{bow}_{G}|}{|\text{bow}_{E}|} == 1]
\end{equation}

The vectors’ vocabulary is limited to a domain determined by the question type. Questions are classified into domains such as yes/no, color, shape, or number, and each domain uses a predefined vocabulary list (see \autoref{fig:categories}).

\begin{figure*}
\begin{lstlisting}
yesno_set = {'yes', 'no'}
color_set = {
    'orangebrown', 'spot', 'yellow', 'blue', 'rainbow', 'ivory', 
    'brown', 'gray', 'teal', 'bluewhite', 'orangepurple', 'black', 
    'white', 'gold', 'redorange', 'pink', 'blonde', 'tan', 'turquoise', 
    'grey', 'beige', 'golden', 'orange', 'bronze', 'maroon', 'purple', 
    'bluere', 'red', 'rust', 'violet', 'transparent', 'yes', 'silver', 
    'chrome', 'green', 'aqua'
}
shape_set = {
    'globular', 'octogon', 'ring', 'hoop', 'octagon', 'concave', 'flat', 
    'wavy', 'shamrock', 'cross', 'cylinder', 'cylindrical', 'pentagon', 
    'point', 'pyramidal', 'crescent', 'rectangular', 'hook', 'tube', 
    'cone', 'bell', 'spiral', 'ball', 'convex', 'square', 'arch', 'h', 
    'cuboid', 'step', 'rectangle', 'dot', 'oval', 'circle', 'star', 
    'crosse', 'crest', 'octagonal', 'cube', 'triangle', 'semicircle', 
    'domeshape', 'obelisk', 'corkscrew', 'curve', 'circular', 'xs', 
    'slope', 'pyramid', 'round', 'bow', 'straight', 'triangular', 
    'heart', 'fork', 'teardrop', 'fold', 'curl', 'spherical', 
    'diamond', 'keyhole', 'conical', 'dome', 'sphere', 'bellshaped', 
    'rounded', 'hexagon', 'flower', 'globe', 'torus'
}   
\end{lstlisting}
\caption{Keywords for WebQA question categories.}
\label{fig:categories}
\end{figure*}

\end{document}